\documentclass[10pt,twocolumn,letterpaper]{article}

\usepackage{iccv}
\usepackage{times}
\usepackage{epsfig}
\usepackage{graphicx}
\usepackage{amsmath}
\usepackage{amssymb}

\usepackage{algorithm}
\usepackage{algorithmic}

\usepackage{setspace}
\usepackage{multirow}

\usepackage{graphicx}

\usepackage{tikz}
\usepackage{tikz-3dplot}
\usetikzlibrary{fadings}

\usepackage{booktabs}

\usepackage{array}
\usepackage{xcolor}
\usepackage{colortbl}

\usepackage{booktabs}

\usepackage[pagebackref=true,breaklinks=true,letterpaper=true,colorlinks,bookmarks=false]{hyperref}

\newcommand{\cgr}{\cellcolor[gray]{0.9}}

\iccvfinalcopy % *** Uncomment this line for the final submission

\def\iccvPaperID{****} % *** Enter the ICCV Paper ID here

\ificcvfinal\fi

\begin{document}
%%%%%%%%% TITLE
\title{SCoTTi: Save Computation at Training Time with an adaptive framework}

\author{Ziyu Li\qquad Enzo Tartaglione\qquad Van-Tam Nguyen\\
LTCI, Télécom Paris, Institut Polytechnique de Paris, France\\
{\tt\small \{ziyu.li, enzo.tartaglione, van-tam.nguyen\}@telecom-paris.fr}
}

\maketitle
\thispagestyle{empty}

%%%%%%%%% ABSTRACT start
\begin{abstract}
    On-device training is an emerging approach in machine learning where models are trained on edge devices, aiming to enhance privacy protection and real-time performance. However, edge devices typically possess restricted computational power and resources, making it challenging to perform computationally intensive model training tasks. Consequently, reducing resource consumption during training has become a pressing concern in this field. To this end, we propose SCoTTi (Save Computation at Training Time), an adaptive framework that addresses the aforementioned challenge. It leverages an optimizable threshold parameter to effectively reduce the number of neuron updates during training which corresponds to a decrease in memory and computation footprint. Our proposed approach demonstrates superior performance compared to the state-of-the-art methods regarding computational resource savings on various commonly employed benchmarks and popular architectures, including ResNets, MobileNet, and Swin-T.\footnote{This article has been accepted for publication at the Resource Efficient Deep Learning for Computer Vision Workshop (ICCV 2023 workshop).}
\end{abstract}
%%%%%%%%% ABSTRACT end

\section{Introduction}
In recent years, there has been a growing interest in the development and deployment of machine-learning models due to their exceptional performance in various domains. However, the remarkable success of these models has come at a cost, as they require substantial computational resources for training and deployment, which makes training on resource-constrained edge devices very difficult~\cite{DBLP:journals/corr/abs-2007-05558}. To this end, two distinct research directions have emerged, each focusing on a different aspect of model optimization.

\begin{figure}[t]
\begin{center}
   \includegraphics[width=0.9\linewidth]{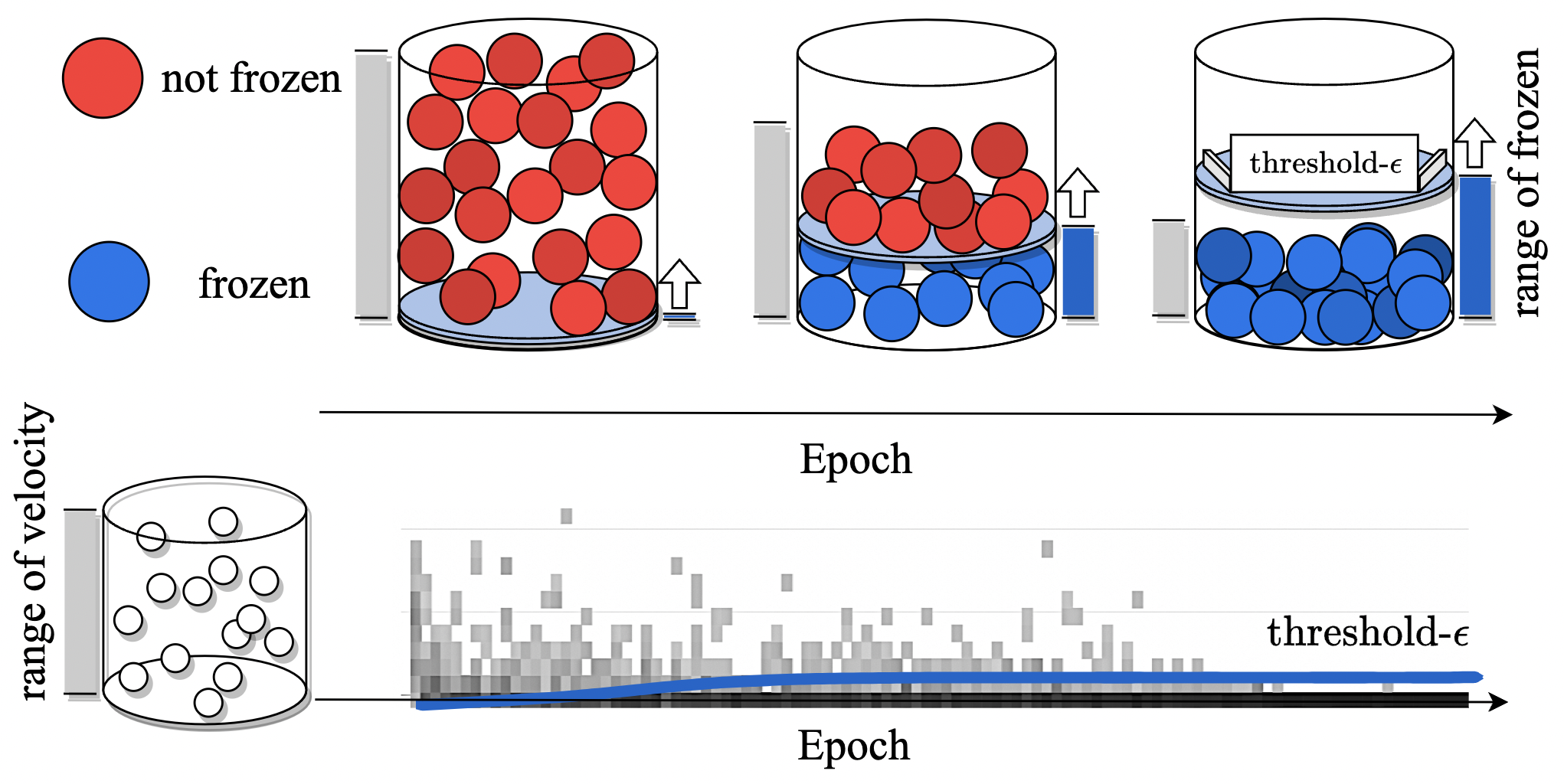}
\end{center}
   \caption{Our adaptive framework SCoTTi is based on an optimizable threshold $\epsilon$. During the training, some neurons learn their own function (their change rate is measured through a velocity): all the neurons whose velocity is lower than some threshold $\epsilon$ will be stopped from updating.}
\label{fig: method}
\end{figure}

The first line of research centers around pruning algorithms, which aim to reduce the size of machine learning models without significantly compromising their performance~\cite{NIPS1989_6c9882bb}. Pruning algorithms selectively remove redundant or less influential components of a model, such as connections or parameters, while preserving its overall functionality~\cite{NIPS2015_ae0eb3ee, Li_2022_CVPR}. By doing so, these algorithms target the reduction of the memory footprint and computational requirements of the model without sacrificing its accuracy~\cite{radford2018improving}.

In parallel, researchers have also been exploring ways to save computational resources during the training process itself~\cite{jiao2018energy,servais2021adaptive,yang2023efficient}. Training deep learning models typically involves numerous iterations over vast amounts of data, which can be computationally expensive and time-consuming, especially when moving to microcontrollers in the context of TinyML models~\cite{pau2023suitability, 10207930}. Therefore, finding methods to reduce the computational burden without compromising the training quality has become an important research objective~\cite{huang2020loadaboost}, also considering that some steps are being moved to design hardware supporting dynamic pruning~\cite{wang2022trainer}. 
The lottery hypothesis~\cite{frankle2018lottery,tartaglione2022rise} highlights that only a small fraction of the tens of thousands of neurons in a neural network significantly impact the model's performance. To save resources, we only need to update the portion of neurons that play a key role.

In our research, we updated only a subset of neurons during backpropagation and showed its feasibility through experiments. One approach is to optimize hyper-parameters like the learning rate through gradient descent~\cite{chandra2022gradient}, to ensure dynamic and faster convergence across the learning. In most instances, a trade-off between accuracy and saved Floating Point Operations (FLOPs) is inevitable, as we will demonstrate in Tab.~\ref{tab:results} (Sec.~\ref{sec: Experimental Results}). To address this issue, we propose SCoTTi, a framework that allows dynamic selection of the neurons requiring an update at the current optimization step. More specifically, we leverage over the concept of \emph{neuron velocity} (introduced in~\cite{bragagnolo2022update}) which is an estimator of learning convergence at the single neuron scale, to determine the sub-network to be updated. We propose a scheme where the velocity threshold $\epsilon$, which determines whether neurons having a certain (low) velocity are in a frozen state or not, is learned at training time. Fig.~\ref{fig: method} pictures the practical effect of SCoTTi: $\epsilon$ is naturally, through stochastic gradient descent, naturally increased as the average velocities (in the distribution at the bottom) in average drop to zero. SCoTTi has the big advantage of not requiring relevant hyper-parameters tuning (as both the learning rate and $\epsilon$ are learned), which ensures flexibility to the specific task and at the same time efficiency in terms of FLOPs computation at training time and to the best of our knowledge, is the first approach proposing a dynamic learned threshold to determine which part of the network to be learned.

The rest of the paper is organized as follows. In Sec.~\ref{sec:related} we review the relevant literature concerning both pruning for deep neural networks and efficient hyper-parameters tuning. Next, in Sec.~\ref{sec:method} we describe SCoTTi, first introducing the working mechanisms for the ultimate optimizer Sec.~\ref{sec: Ultimate Optimizer} and of neurons at equilibrium Sec.~\ref{sec: Neurons at Equilibrium}, and then describing the method proposing as well a rationale behind its working principle. Then, in Sec.~\ref{sec:exp}  we experiment with our proposed training scheme over some deep ANNs on many different datasets, and finally, Sec.~\ref{sec:conclusion} concludes while providing further directions for future research.

\section{Related Works}
\label{sec:related}

In this section, we will provide an overview of some of the most popular approaches to save computation when employing deep neural networks. To this end, we will divide the most-relevant literature into pruning-based approaches, distributed learning, and the most recent new directions to efficient parameters update.

\subsection{Pruning methods}
Pruning and quantization have emerged as prominent model compression techniques and have witnessed extensive development and application in recent years. Three mainstream pruning strategies - weight pruning~\cite{NIPS1989_6c9882bb}, filter pruning~\cite{DBLP:journals/corr/LiKDSG16}, and neurons pruning~\cite{DBLP:journals/corr/abs-2111-08577} - have rapidly become hotspots of research. Weight pruning achieves model sparsity by removing small-magnitude weights in deep neural networks, thereby reducing the model's parameter count and computational complexity~\cite{DBLP:journals/corr/abs-2006-02768, DBLP:journals/corr/abs-1804-03294}. Filter pruning focuses on entire convolutional filters (channels), eliminating unimportant filters to further compress the model~\cite{DBLP:journals/corr/abs-1911-07412}. Neurons pruning simplifies the model by removing entire neurons that make minor contributions to the model's output. Additionally, model quantization further reduces model size and computational requirements by representing model parameters with lower precision, which is especially beneficial for edge devices with limited memory and computation capabilities~\cite{DBLP:journals/corr/abs-2106-08295, DBLP:journals/corr/abs-1911-09464}.
The application of these pruning and quantization methods significantly enhances model efficiency on edge devices, enabling the deployment of more complex models and achieving outstanding performance in real-time applications. Despite the big interest of the community, pruning, and quantization methods also exhibit notable limitations. Firstly, pruning requires multiple iterations to remove model parameters, which entails significant computation and time consumption~\cite{tartaglione2020pruning}. Quantization, on the other hand, may result in some loss of model accuracy due to reduced precision~\cite{zhou2018explicit}. Additionally, while these techniques effectively optimize resource consumption during model inference on edge devices, they offer limited assistance in training new models and require specific techniques to address this issue~\cite{liu2021transtailor}. Consequently, retraining models on edge devices becomes a challenging task, calling for innovative solutions to overcome these restrictions.

\subsection{Distributed learning} 
So far, most methods for training on edge devices are based on the idea of distributed learning~\cite{DBLP:journals/corr/abs-1912-09789}. These methods allow model training on edge devices without uploading raw data to a central server. On-device training methods based on distributed learning or federated learning offer significant advantages, including enhanced privacy protection as training occurs locally on edge devices without requiring data uploads to central servers~\cite{lu2022privacypreserving, balcan2012distributed}. This approach reduces data transmission costs and alleviates the burden on central servers, contributing to lower inference latency. Moreover, the near real-time model updates make these methods suitable for time-sensitive applications.

Distributed learning or federated learning-based method has, however, certain limitations. The limited computational and storage resources of edge devices can lead to slower training efficiency and a prolonged convergence time~\cite{e22050544}. Communication overhead remains a concern, as model updates among devices might result in substantial network traffic, particularly in scenarios with a large number of devices~\cite{inproceedings}. The heterogeneity of edge devices poses challenges, potentially affecting training stability and convergence~\cite{app12189124}. Furthermore, the complexity of federated learning algorithms requires careful consideration of privacy preservation, model aggregation, and model drift, making algorithm design and optimization demanding tasks~\cite{CAO2022102413}.

\subsection{Efficient parameters update schemes} 
The lottery ticket hypothesis~\cite{frankle2018lottery} posits that within a neural network, only a small subset of neurons plays a crucial role in achieving high accuracy, while the majority of neurons are less significant. Based on this hypothesis, if we focus on updating only this critical subset of neurons during the training process, we can attain comparable accuracy to updating all the neurons.
The advantage of this approach is that it leads to more efficient training and improved generalization performance. By updating only the essential neurons, the training process focuses on the most informative parts of the model, mitigating the negative influence of noise and less relevant parameters. The challenge to identify such sub-networks efficiently is, however, still ongoing~\cite{tartaglione2022rise}. Some first steps have been moved in such a direction, achieving similar accuracy to a fully trained model while requiring fewer iterations and computational resources~\cite{bragagnolo2022update}. SCoTTi puts itself in such a frame, building on top of recent advances in the field~\cite{bragagnolo2022update, chandra2022gradient} and providing a simple yet effective approach that automatically tunes the updated sub-graph at training time, leading to computation savings at training time. In the next section, we will ground and provide more details on the working principles of SCoTTi.

\section{SCoTTi}
\label{sec:method}
\begin{figure*}[ht]
\begin{center}
   \includegraphics[width=1.0\linewidth]{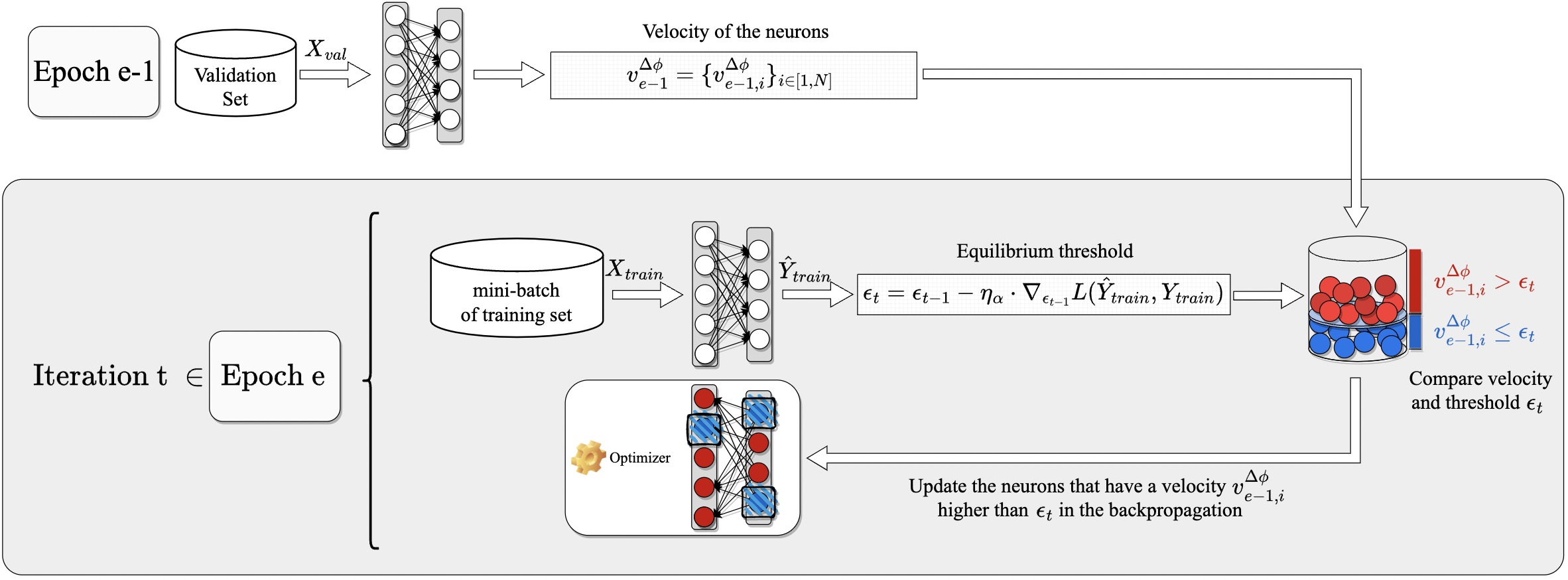}
\end{center}
   \caption{In the given figure, $X_{train}$ and $X_{val}$ correspond to the input of the training set $\Xi_{train}$ and the validation set $\Xi_{val}$, respectively. $Y_{train}$ corresponds to the labels of the training set. $\hat{Y}_{train}$ represents the outputs obtained from the model when the corresponding inputs from the training set are fed into it. We update the threshold $\epsilon$ by the model's loss on the training set and compare it to the velocity of each neuron, suppressing updates to neurons with a velocity less than $\epsilon$, where velocity is a value that indicates the similarity between the outputs of the same neuron in two consecutive epochs.}
\label{fig: process}
\end{figure*}
In this section, we will describe our proposed approach, which is pictured in Fig.~\ref{fig: process}. It builds on top of two approaches: \emph{Gradient descent the ultimate optimizer}~\cite{DBLP:journals/corr/BaydinCRSW17} (reviewed in Sec.~\ref{sec: Ultimate Optimizer}) and \emph{To update or not to update? Neurons at equilibrium in deep models }~\cite{bragagnolo2022update} (reviewed in Sec.~\ref{sec: Neurons at Equilibrium}). The innovation of SCoTTi is to make these algorithms work synergetically and optimize over a (formerly) static hyper-parameter ($\epsilon$) that determines a threshold above which a neuron should be kept updated. This hyper-parameter is optimized along the training process and hence does not require fine optimization (as described in Sec.~\ref{sec:epsilon}).

\subsection{Learning the learning rate}
\label{sec: Ultimate Optimizer}
The optimization of Deep Neural Networks (DNN) is a critical task that requires finding suitable hyperparameters (eg. learning rate) for the optimizer. 
% 深度神经网络（DNN）的优化是一项关键任务，需要为优化器找到合适的超参数（如学习率）。
The optimization of hyperparameters has been introduced in a previous work \cite{DBLP:journals/corr/BaydinCRSW17}; however, this method suffers from three main limitations. Firstly, the manual differentiation of optimizer update rules is a laborious and error-prone process that needs to be repeated for each variant of the optimizer. Secondly, the method only focuses on tuning the learning rate and neglects other important hyperparameters such as the momentum coefficient. Lastly, the method introduces an additional hyperparameter, the hyper-learning rate, which also requires tuning. These limitations highlight the need for an improved approach to address these challenges in hyperparameters optimization.

The ultimate optimizer~\cite{chandra2022gradient} is a recently proposed optimization scheme that operates based on the principle of dynamically updating the original fixed learning rate and/or other hyperparameters in real-time. It is designed to complement and enhance the functionality of general optimizers such as Adam~\cite{kingma2017adam}, SGD~\cite{Bottou2010LargeScaleML}, and others. The primary objective of the ultimate optimizer is to adaptively adjust the learning rate and/or other hyperparameters throughout the training process to find the most appropriate and effective one/ones for the current state of the model. By dynamically updating the hyperparameters like the learning rate, the ultimate optimizer aims to improve the convergence speed, optimization efficiency, and overall performance of the neural network model:

\begin{align}
 \alpha_{t+1}&=\alpha_t-\eta \cdot \dfrac{\partial f(\omega_t)}{\partial \alpha_t} \label{eq: update hyper}\\
\omega_{t+1}&=\omega_t- \alpha_{t+1} \cdot \dfrac{\partial f(\omega_t)}{\partial \omega_t}. \label{eq: update original}
\end{align}
As shown in \eqref{eq: update original}, the ultimate optimizer modified the traditional approach of updating weight $\omega$ by treating the learning rate $\alpha$ as a learnable parameter, as outlined in \eqref{eq: update hyper}, the approach of updating learning rate $\alpha$ which is based on the gradient descent algorithm, an additional hyperparameter $\eta$ is required to control the step size for updating the learning rate $\alpha$ through gradient descent.

%-------------------------------------------------------------------------
\subsection{Neurons at Equilibrium}
\label{sec: Neurons at Equilibrium}

It is widely recognized in the literature that Deep Learning models commonly employed in state-of-the-art scenarios tend to be over-parametrized~\cite{DBLP:journals/corr/abs-1710-10174}. This observation has spurred two lines of research: reducing the size of these models through pruning algorithms, or finding ways to save computational resources during the training process. The former approach has received significant attention and has been extensively investigated. However, the latter approach had been at an impasse until a recent study \cite{frankle2018lottery} suggested its viability.

As shown in the lottery ticket hypothesis, only a small fraction of the total number of neurons in the model can be considered effective or influential in determining the model's performance (ie. The model's performance heavily relies on a subset of neurons that play a decisive role in capturing and representing the relevant patterns and information in the given data.). As inferred by the authors of Neurons at Equilibrium (NEq)~\cite{bragagnolo2022update}, the existence of redundancy or inefficiency in the training process, where a large number of neurons may get updates despite not playing a decisive role in capturing the underlying patterns or information in the data. By identifying and mitigating these unnecessary updates, it is possible to optimize the training process and improve the overall efficiency of the model. To this end, we want to identify the crucial neurons that require updating in each epoch based on a mask that is generated by employing a fixed threshold-$\epsilon$. 

\begin{figure}
\begin{center}
   \includegraphics[width=1.0\linewidth]{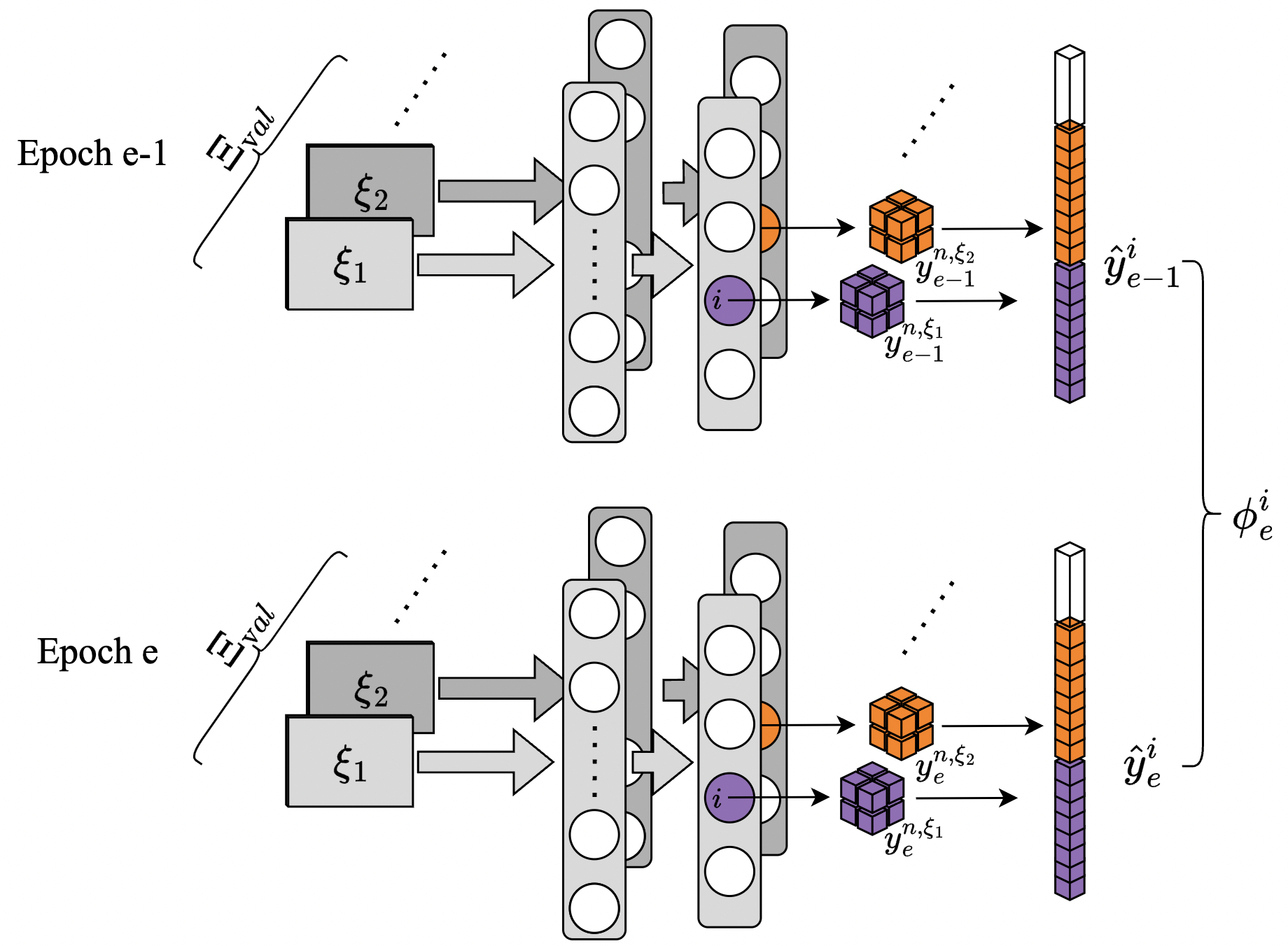}
\end{center}
   \caption{For a given epoch $e$, the model receives the $\xi$-th sample from the validation set $\Xi_{val}$. The output of the $i$-th neuron depends on both the model’s parameters and the specific sample on the validation set. These outputs are squeezed, concatenated and the obtained vector is then normalized, obtaining $\hat{y}^i_e$. And we denote that $\phi_e^i$ is the cosine similarity of the feature vector $\hat{y}^i_e$ and the feature vector $\hat{y}^i_{e-1}$ which is obtained at epoch $e-1$.}
\label{fig: NEq}
\end{figure}

The method for determining whether a neuron requires a further update or not follows.
First, the cosine similarity between the outputs of the same neuron in two consecutive epochs is calculated below and also shown in Fig.~\ref{fig: NEq}:

\begin{align}
\phi_e = \cos(\theta) 
% &= \frac{\hat{y}_{e-1} \cdot \hat{y}_e}{\lVert \hat{y}_{e-1} \rVert \cdot \rVert \hat{y}_{e} \rVert} \\[3pt]
= \hat{y}_{e-1} \cdot \hat{y}_e 
= \sum_{\xi\in\Xi_{val}}\sum^{N}_{n=1}\hat{y}^{n,\xi}_{e}\cdot \hat{y}^{n,\xi}_{e-1} .
\end{align}

We further use the obtained similarity to calculate the corresponding velocity as shown below:
\begin{align}
\Delta\phi_e &=\phi_e-\phi_{e-1}\\[3pt]
v_e &=\Delta\phi_e-\mu_{eq}\cdot v_{e-1} .
\end{align}
This is then compared to some threshold $\epsilon$ to determine if the neuron should be updated. The condition determining whether a neuron is in a frozen state is: 

\begin{equation}
|v_e| \le \epsilon .\label{eq: condition1}
\end{equation}

When the inequality \eqref{eq: condition1} is satisfied, it indicates that the neuron has reached an equilibrium state at epoch $e$.\footnote{The state of a neuron at epoch $e+1$ depends on two factors: its parameters and the working region (domain). When frozen, while the parameters do not change, the working region might (unless all the neurons of the previous layers are frozen). Hence, the state of a neuron might unfreeze.} Consequently, the neuron will be detached from the back-propagation graph, indicating that it should not be updated at epoch $e$. The effectiveness of this approach depends on two main factors: the optimality of the learning schedule, and the optimal, dynamic tuning of $\epsilon$. While the first is addressed in Sec.~\ref{sec: Ultimate Optimizer}, in the next section we will propose an approach to dynamically learn $\epsilon$.

\subsection{Learning the velocity threshold}
\label{sec:epsilon}
In our study, we employ the NEq method in conjunction with the ultimate optimizer to update the weights. Additionally, we introduce the threshold parameter $\epsilon$ as an additional optimizable parameter to the update paradigm. After rearranging the terms of \eqref{eq: condition1}, the previous inequality can be expressed as follows:

\begin{equation}
|v_e| - \epsilon \le 0 .\label{eq: condition2}
\end{equation}

We define the $i$-th neuron to be in the equilibrium state at epoch $e$ if \eqref{eq: condition2} is satisfied.
% \iffalse
% \fi
Our approach is currently based on the SGD optimizer (but extendible to any other optimizer), therefore to incorporate $\epsilon$ as a learnable parameter that can be optimized by the hyper optimizer as shown in Fig.~\ref{fig: process}, we integrate the inequality into the weight update paradigm, as illustrated below. We utilize step function $\Theta(\cdot)$ to serve as the activation function for this component:

\begin{equation}
    \omega_{t}=\omega_{t-1}-\alpha\cdot\dfrac{\partial \mathcal{L}}{\partial \omega_t}{\color{black}\cdot \Theta(|v_e| - \epsilon) }. \label{eq: update}
\end{equation}

As shown in \eqref{eq: update}, $\alpha$ represents the learned learning rate, $\epsilon$ presents the learned threshold, $\omega_{t}$ represents the weight update at epoch t, $\frac{\partial \mathcal{L}}{\partial \omega_t}$ refers to the gradient of the loss function, and $\Theta(\cdot)$ is the step function that ensures the inequality condition is met. The hyperparameters $\alpha$ and $\epsilon$ are iteratively updated by another hyper optimizer, following the gradient of the loss function for each parameter, denoted as $\frac{\partial \mathcal{L}}{\partial \alpha}$ and $\frac{\partial \mathcal{L}}{\partial \epsilon}$, respectively. This update scheme ensures that the parameters are adjusted in a direction that minimizes the loss function, thereby leading to a reduction in overall loss.
By incorporating the $\Theta(\cdot)$ term, compared to the original update paradigm, our approach presents two major advantages.

\paragraph{We introduce $\epsilon$ as a learnable parameter} The inclusion of $\Theta(\cdot)$ enables us to incorporate $\epsilon$ into the update paradigm as an optimizable parameter. Specifically, $\epsilon$ is optimized along the direction of loss reduction, ensuring that updates are made in a manner that progressively minimizes the overall loss. This allows us to dynamically adjust the threshold for determining neuron equilibrium during the optimization process.

\paragraph{SCoTTi is consistent with ``freezing neurons at equilibrium''} The use of the $\Theta(\cdot)$ function ensures that if a neuron is in equilibrium, $\Theta(\cdot)$ will filter these ones. Conversely, if the neuron is not in equilibrium, $\Theta(\cdot)$ will allow updates. This aligns with the concept of NEq, where only non-equilibrium neurons undergo updates while equilibrium neurons remain unchanged.

In the following, we will show how to update $\alpha$ and $\epsilon$ by a parameter $\eta$ via gradient descent per iteration and derive an iterative update paradigm, being for $\epsilon$.

\begin{equation}
\begin{aligned}
\epsilon_{t+1}&=\epsilon_{t}-\eta\cdot\dfrac{\partial \mathcal{L}(w_t)}{\partial \epsilon_t} =\epsilon_{t}-\eta\cdot\dfrac{\partial \mathcal{L}(w_t)}{\partial w_t}\cdot\dfrac{\partial w_t}{\partial \alpha_t}\cdot\\
&=~\epsilon_{t}-\eta\cdot\dfrac{\partial \mathcal{L}(w_t)}{\partial w_t}\\
&\qquad\cdot\dfrac{\partial \left [w_{t-1}-\alpha_t\cdot\dfrac{\partial \mathcal{L}(w_{t-1})}{\partial w_{t-1}}\cdot \Theta(|v_e|-\epsilon_t)\right ]}{\partial \epsilon_t}\\[3pt]
&=\epsilon_{t}-\eta\cdot {\dfrac{\partial \mathcal{L}(w_t)}{\partial w_t}\cdot\dfrac{\partial \mathcal{L}(w_{t-1})}{\partial w_{t-1}}}\cdot{\color{black}\alpha_t},
\end{aligned}
\end{equation}

\noindent while for $\alpha$

\begin{equation}
\begin{aligned}
\alpha_{t+1}&=\alpha_{t}-\eta\cdot\dfrac{\partial \mathcal{L}(w_t)}{\partial \alpha_t} =\alpha_{t}-\eta\cdot\dfrac{\partial \mathcal{L}(w_t)}{\partial w_t}\cdot\dfrac{\partial w_t}{\partial \alpha_t}\\[3pt]
&=\alpha_{t}-\eta\cdot\dfrac{\partial \mathcal{L}(w_t)}{\partial w_t}\\[3pt]
&\qquad\cdot\dfrac{\partial \left [w_{t-1}-\alpha_t\cdot\dfrac{\partial \mathcal{L}(w_{t-1})}{\partial w_{t-1}}\cdot \Theta(|v_e|-\epsilon_{t})\right ]}{\partial \alpha_t}\\[3pt]
&=\alpha_{t}+\eta\cdot{\dfrac{\partial \mathcal{L}(w_t)}{\partial w_t}\cdot\dfrac{\partial \mathcal{L}(w_{t-1})}{\partial w_{t-1}}}\cdot \Theta(|v_e|-\epsilon_{t}).
\end{aligned}
\end{equation}

\noindent Indeed, the iterative update paradigm enables us to optimize both $\alpha$ and $\epsilon$ simultaneously. The iterative process continues until the values of $\alpha$ and $\epsilon$ converge to their optimal values, indicating that the model has reached an optimal state.

\begin{algorithm}[t]
    \caption{Learned threshold $\epsilon$ approach}
    \begin{algorithmic}[1]
    \setstretch{1.1}
        \REQUIRE{\qquad
        \\$\mathcal{L}_{t-1}$: loss calculated on the current minibatch,
        \\$\boldsymbol{\mathcal{E}}$: a dictionary used to store the neurons that are at equilibrium state,
        \\$\boldsymbol{v}_e$: a dictionary used to record the velocity between epochs $e-1$ and $e$,
        \\$\boldsymbol{\omega}_{t-1}$: parameters of the model at iteration $t-1$,
        \\$\alpha_{t-1}$: learnable learning rate, at iteration $t-1$,
        \\$\epsilon_{t-1}$: learnable threshold, at iteration $t-1$,
        \\$\eta_\alpha$: hyper-learning rate for $\alpha$,
        \\$\eta_\epsilon$: hyper-learning rate for $\epsilon$.}
        \\~\\
        \STATE $\alpha_{t} \gets \alpha_{t-1} - \eta_\alpha\cdot\nabla_{\alpha_{t-1}}\mathcal{L}_{t-1}$\label{line:hypLR}
        \STATE $\epsilon_{t} \gets \epsilon_{t-1} - \eta_\epsilon\cdot\nabla_{\epsilon_{t-1}}\mathcal{L}_{t-1}$\label{line:hypeps}
        \FOR {all the $i$-th neurons in the model}
        
            \IF{i is in $\boldsymbol{\mathcal{E}}$}\label{line:skip}
            \STATE $G^i_{t-1} = 0$
            \ELSE
            \STATE $G^i_{t-1} = \nabla_{\omega_{t-1}^i}\mathcal{L}_{t-1}$
            \ENDIF

            \STATE $\omega_{t}^i \gets \omega_{t-1}^i-\alpha_t\cdot G^i_{t-1} \cdot \Theta(|v_e^{i}|-\epsilon_t)$\label{line:update}
        
        \ENDFOR
        \RETURN $\alpha_t$, $\epsilon_t$, $\boldsymbol{\omega}_{t}$
    \end{algorithmic}
    \label{algo: a2}
\end{algorithm}

\subsection{Overview on the method}
As presented in Algorithm~\ref{algo: a2}, we incorporate a mask in the ultimate optimizer to constrain the weight updates for a part of neurons (in the form of a dictionary, $\boldsymbol{\mathcal{E}}$). Additionally, we introduce an additional parameter $\epsilon$, into the standard iterative update paradigm for the weights. This enables us to learn the threshold $\epsilon$ for neuron equilibrium and control the updates accordingly in real time. Focusing on the hyper-optimization step, we extend the hyper-optimizer for the learning rate (line~\ref{line:hypLR}) to include updates for $\epsilon$ (line~\ref{line:hypeps}). By optimizing both of them simultaneously, we can fine-tune the threshold for determining neuron equilibrium and further enhance the model's performance. When performing the update in line~\ref{line:hypeps}, to maintain the gradient flow without disruption, we employ a Straight-Through Estimator (STE)~\cite{DBLP:journals/corr/BengioLC13}. STE allows for backpropagation through the weight update operation without modifying the gradient. This technique enables us to incorporate non-differentiable operations, such as binarization which we used here (neurons in a frozen state), into the training process while preserving the gradient information. We can successfully skip the gradient computation for the neurons marked in a frozen state (line~\ref{line:skip}) and successively we perform a standard update for the model's parameters (line~\ref{line:update}).

\section{Experiment}
\label{sec:exp}

In this section, we present our experiments describing datasets, architectures, and learning policies. We first propose an ablation study (Sec.~\ref{sec: Ablation Study}) and then we describe the main experimental results (Sec.~\ref{sec: Experimental Results}). We have performed our experiments on an NVIDIA~Tesla~V100 equipped with 32GB and developed the code using PyTorch~1.13.1.\footnote{The code is publicly available at \href{https://github.com/liziyu403/SCoTTi-Save-Computation-at-Training-Time-with-an-adaptive-framework}{https://github.com/liziyu403/SCoTTi
\\-Save-Computation-at-Training-Time-with-an-adaptive-framework}.}

\begin{table*}[t]
\caption{Table of experimental results. \textbf{FLOPs saved} refers to the total FLOPs saved at the end of model training as a percentage of the total FLOPs when the model is updated normally (no frozen); \textbf{Top-1} refers to the model's Test Top1-Accuracy. Models marked with ``*'' indicate that they utilize a pre-trained model on ImageNet1k. The rows in gray are our proposed SCoTTi.}
%\label{tab:results}
\centering

\label{tab:results}
\resizebox{\textwidth}{!}{
    \begin{tabular}{ccccccc}
        \toprule
        \multirow{2}{*}{\textbf{Dataset}} & \multirow{2}{*}{\textbf{Architecture}} & \multicolumn{3}{c}{\textbf{Optimization Approach} } & \multirow{2}{*}{\textbf{FLOPs saved}} & \multirow{2}{*}{\textbf{Top-1}} \\
        \cline{3-5}
                        &               & NEq~\cite{bragagnolo2022update} & Ultimate optim.~\cite{chandra2022gradient} & Learned $\epsilon$\\
        
        %%%%%%%%%% CIFAR10 result %%%%%%%%%%
        \midrule
        \multirow{10}{*}{CIFAR-10 \cite{Krizhevsky2009LearningML}} 
        & \multirow{5}{*}{VGG-16 \cite{7486599}} 
        & \qquad\qquad\qquad %scheduler SGD ( \texttt{baseline} ) 
        &&& $00.00\%$ & $88.54\%$ \\

        && %scheduler SGD with NEq 
        \checkmark&&& $37.41\%$ & $89.86\%$ \\
        && %Ultimate Optimizer 
        &\checkmark&& $00.00\%$ & \textbf{92.76\%} \\

        && %Ultimate Optimizer with NEq 
        \checkmark&\checkmark&& $30.98\%$ & $92.70\%$ \\

        && %Our Learned $\epsilon$ Approach 
        \cgr  \checkmark&\cgr\checkmark&\cgr\checkmark&\cgr \textbf{43.66\%} &\cgr  92.58\%\\

        \cmidrule{2-7} 
        
        & \multirow{5}{*}{Swin-T$^{*}$ \cite{DBLP:journals/corr/abs-2103-14030}} 

        & \qquad\qquad\qquad %scheduler SGD ( \texttt{baseline} ) 
        &&& $00.00\%$ & $91.59\%$ \\

        && %scheduler SGD with NEq 
        \checkmark&&& $39.66\%$ & $90.96\%$ \\

        && %Ultimate Optimizer 
        &\checkmark&& $00.00\%$ & $91.65\%$ \\

        && %Ultimate Optimizer with NEq 
        \checkmark&\checkmark&& $48.84\%$ & $91.74\%$ \\

        && %Our Learned $\epsilon$ Approach 
        \cgr \checkmark&\cgr\checkmark&\cgr\checkmark&\cgr \textbf{58.76\%} &\cgr  \textbf{91.77\%}\\
        %%%%%%%%%% CIFAR100 result %%%%%%%%%%
        \midrule
        \multirow{10}{*}{CIFAR-100 \cite{Krizhevsky2009LearningML}} 
        & \multirow{5}{*}{ResNet-32 \cite{7780459}} 

        & \qquad\qquad\qquad %scheduler SGD ( \texttt{baseline} ) 
        &&& $00.00\%$ & $68.42\%$ \\

        && %scheduler SGD with NEq 
        \checkmark&&& $38.80\%$ & $69.97\%$ \\

        && %Ultimate Optimizer 
        &\checkmark&& $00.00\%$ & $70.06\%$ \\

        && %Ultimate Optimizer with NEq 
        \checkmark&\checkmark&& $59.89\%$ & $69.24\%$ \\

        && %Our Learned $\epsilon$ Approach 
        \cgr  \checkmark&\cgr\checkmark&\cgr\checkmark&\cgr \textbf{60.59\%} &\cgr  \textbf{70.43\%}\\

        \cmidrule{2-7} 

        & \multirow{5}{*}{ResNet-56 \cite{7780459}} 

        & \qquad\qquad\qquad %scheduler SGD ( \texttt{baseline} ) 
        &&& $00.00\%$ & $69.69\%$ \\

        && %scheduler SGD with NEq 
        \checkmark&&& $41.12\%$ & $71.40\%$ \\

        && %Ultimate Optimizer 
        &\checkmark&& $00.00\%$ & $70.15\%$ \\

        && %Ultimate Optimizer with NEq 
        \checkmark&\checkmark&& $56.58\%$ & $70.36\%$ \\

        && %Our Learned $\epsilon$ Approach 
        \cgr  \checkmark&\cgr\checkmark&\cgr\checkmark&\cgr \textbf{58.97\%} & \cgr \textbf{71.55\%}\\
        %%%%%%%%%% Domainnet %%%%%%%%%%
        \midrule
        \multirow{5}{*}{Clipart \cite{5206848}} 
        & \multirow{5}{*}{ResNet-18 \cite{7780459}} 

        & \qquad\qquad\qquad %scheduler SGD ( \texttt{baseline} ) 
        &&& $00.00\%$ & $73.21\%$ \\

        && %scheduler SGD with NEq 
        \checkmark&&& $38.06\%$ & $72.19\%$ \\

        && %Ultimate Optimizer 
        &\checkmark&& $00.00\%$ & $73.01\%$ \\

        && %Ultimate Optimizer with NEq 
        \checkmark&\checkmark&& $49.33\%$ & $72.60\%$ \\

        && %Our Learned $\epsilon$ Approach 
        \cgr \checkmark&\cgr\checkmark&\cgr\checkmark&\cgr \textbf{53.86\%} & \cgr \textbf{73.21\%}\\

        %%%%%%%%%% Domainnet %%%%%%%%%%
        %\cline{1-1} \cline{3-7} 
        \midrule
        
        \multirow{5}{*}{Painting \cite{5206848}} 
        & \multirow{5}{*}{ResNet-18 \cite{7780459}} 

        & \qquad\qquad\qquad %scheduler SGD ( \texttt{baseline} ) 
        &&& $00.00\%$ & $64.51\%$ \\

        && %scheduler SGD with NEq 
        \checkmark&&& $27.94\%$ & $62.14\%$ \\

        && %Ultimate Optimizer 
        &\checkmark&& $00.00\%$ & $60.82\%$ \\

        && %Ultimate Optimizer with NEq 
        \checkmark&\checkmark&& \textbf{77.34\%} & $63.46\%$ \\

        && %Our Learned $\epsilon$ Approach 
        \cgr \checkmark&\cgr\checkmark&\cgr\checkmark&\cgr 76.92\% &\cgr  \textbf{65.44\%}\\

        %%%%%%%%%% Tiny ImageNet result %%%%%%%%%%
        \midrule
        \multirow{5}{*}{Tiny ImageNet \cite{5206848}} 
        & \multirow{5}{*}{MobileNet-v2$^{*}$  \cite{DBLP:journals/corr/abs-1801-04381}} 

        & \qquad\qquad\qquad %scheduler SGD ( \texttt{baseline} ) 
        &&& $00.00\%$ & $55.69\%$ \\

        && %scheduler SGD with NEq 
        \checkmark&&& $53.28\%$ & $56.40\%$ \\

        && %Ultimate Optimizer 
        &\checkmark&& $00.00\%$ & $60.02\%$ \\

        && %Ultimate Optimizer with NEq 
        \checkmark&\checkmark&& $80.83\%$ & $60.53\%$ \\

        && %Our Learned $\epsilon$ Approach 
        \cgr  \checkmark&\cgr\checkmark&\cgr\checkmark&\cgr \textbf{86.44\%} &\cgr  \textbf{60.68\%}\\
        \bottomrule
    \end{tabular}
}
\end{table*}

\begin{figure*}[t]
\begin{center}
   \includegraphics[width=1\linewidth]{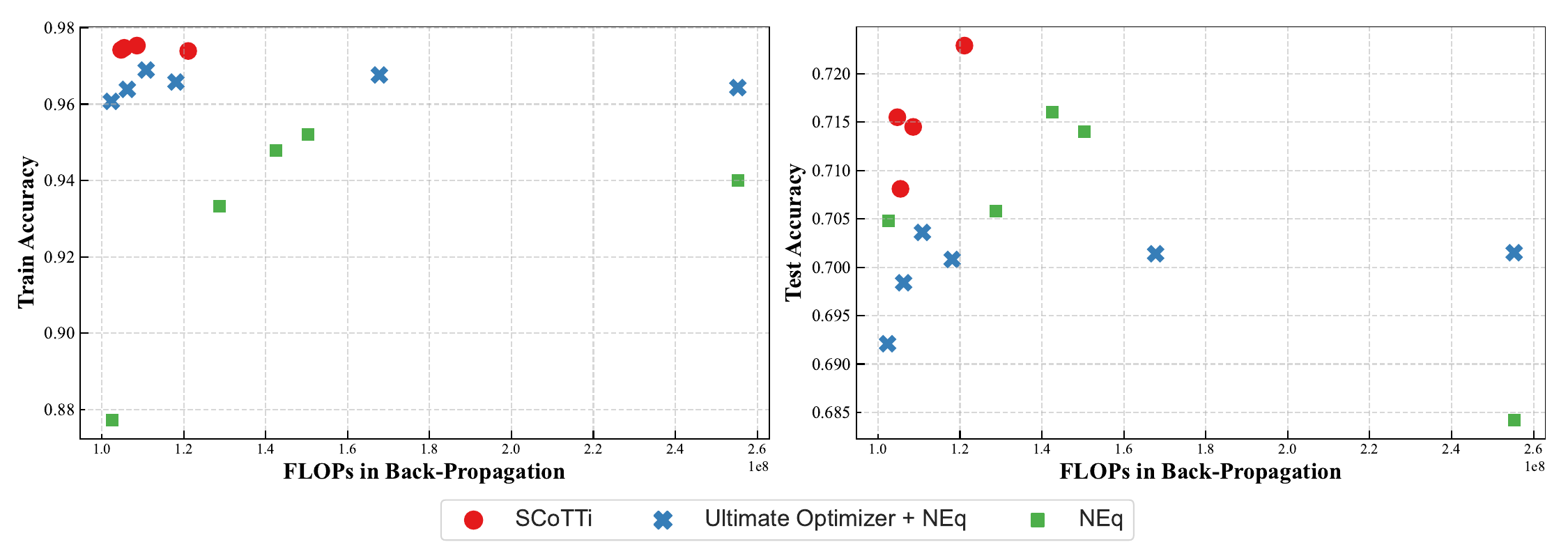}
\end{center}
   \caption{The x-axis represents the average of FLOPs required per epoch, while the y-axis represents the test accuracy achieved. Points closer to the upper left corner indicate better performance, as they represent lower FLOPs and higher accuracy simultaneously.}
\label{fig: compare}
\end{figure*}

\begin{figure}[t]
\begin{center}
   \includegraphics[width=0.8\linewidth]{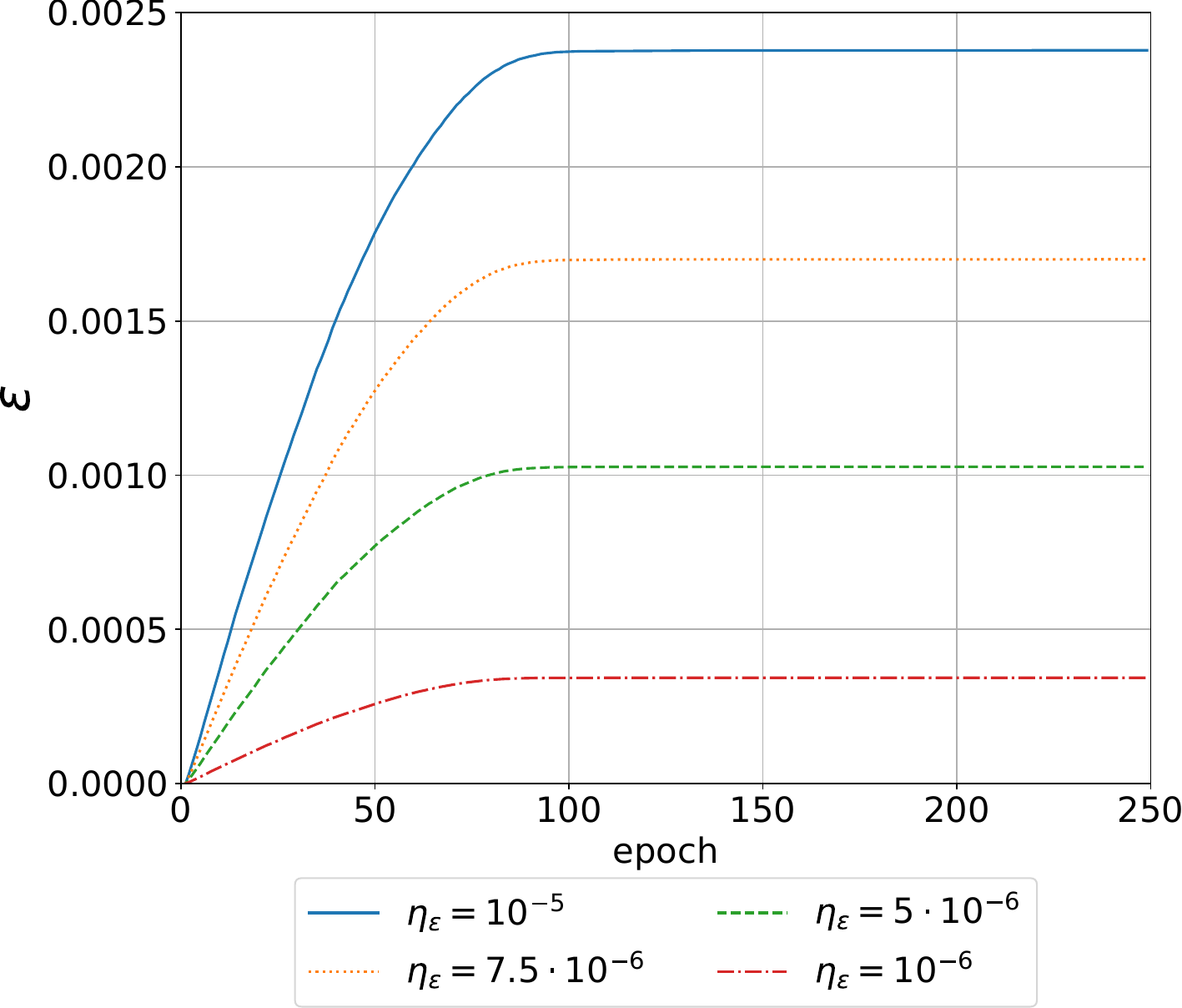}
\end{center}
   \caption{The hyperparameter $\epsilon$ is updated using gradient descent: $\epsilon_{t} = \epsilon_{t-1} - \eta_\epsilon\cdot\nabla_{\epsilon_{t-1}}\mathcal{L}_{t-1}$, with different curves corresponding to different values of $\eta_\epsilon$.}
\label{fig: eps}
\end{figure}

\subsection{Datasets}
\textbf{CIFAR-10~\cite{Krizhevsky2009LearningML}.} The CIFAR-10 dataset is a widely recognized benchmark in image classification. It consists of 60,000 color images divided into 10 classes representing specific object categories, such as airplanes, automobiles, birds, and cats. With 6,000 images per class, the dataset is evenly balanced. The images have a resolution of 32x32 pixels, making them computationally efficient and suitable for scenarios with limited computational resources.

\textbf{CIFAR-100~\cite{Krizhevsky2009LearningML}.} The CIFAR-100 dataset builds upon the CIFAR-10 dataset by introducing 100 fine-grained object categories. It contains 60,000 color images, with 600 images per class. CIFAR-100 offers a higher level of label granularity compared to CIFAR-10, enabling more challenging classification tasks. The images in this dataset also have a resolution of 32x32 pixels. 

\textbf{Tiny ImageNet~\cite{5206848}.} The Tiny ImageNet dataset is more extensive than CIFAR-10 and CIFAR-100, consisting of 200 diverse object categories. It includes a total of 100,000 color images, with 500 images per class. The images in Tiny ImageNet have a resolution of 64x64 pixels, providing a higher level of detail compared to the previous 2 datasets. 

\textbf{Clipart~\cite{peng2019moment}.}
The Clipart dataset contains 73,810 images representing various everyday objects and scenes in a clipart-style visual format. It consists of 345 object categories, making it a rich resource for studying domain adaptation to clipart-like graphics.

\textbf{Painting~\cite{peng2019moment}.}
The Painting dataset consists of 76,174 high-quality images inspired by diverse artistic painting styles. It includes 345 object categories, providing researchers with ample opportunities to study domain adaptation in the context of artistic representations.

\subsection{Ablation study}
\label{sec: Ablation Study}
We performed our ablation study training a ResNet-56 \cite{7780459} on the CIFAR-100 dataset. The baseline models were trained with Stochastic Gradient Descent (SGD) as the optimizer, initialized with a learning rate $\alpha$ of 0.1, momentum $\mu$ of 0.9, and weight decay of 5 × 10$^{-4}$ for a total of 250 epochs. The learning rate $\alpha$ was reduced by a factor of 10 after 100 and 150 epochs. For the experiments based on the Ultimate Optimizer (including SCoTTi), we uniformly set the hyperlearning rate for learning rate to 1.5 × $10^{-5}$. For all other parameters unrelated to the learning rate, we remain consistent with the experiments that use SGD as the optimizer.

In general, a trade-off between FLOPs and accuracy can be observed in Fig.~\ref{fig: compare}: for less than the model's accuracy experiences a significant drop when the average FLOPs value falls below a certain value. In these MFLOPs, we observe a degradation in the performance of SCoTTi. This can be observed by testing several representative $\epsilon$ for each of the two frameworks using a fixed threshold $\epsilon$ (the blue points approach and green points approach are) to determine the relationship between FLOPs of backpropagation and accuracy. For our proposed framework with optimizable threshold $\epsilon$ (in red), we initialize $\epsilon$ to zaro, and since the $\epsilon$ updating process is based on the gradient descent method, a hyperparameter $\eta_\epsilon$ is needed to control the step size when updating the $\epsilon$, and a few representative $\eta_\epsilon$ are selected for recording in our experiments, the learned curves for $\epsilon$ are shown as Fig.~\ref{fig: eps}.

As Fig.~\ref{fig: compare} shows, in these two plots, the top plot represents the results obtained on the TRAIN set, and the bottom plot represents the results on the TEST set. The x-axis represents the average FLOPs required per epoch, while the y-axis represents the accuracy achieved. Our objective is to minimize the FLOPs while maximizing the accuracy. Therefore, points closer to the upper left corner indicate better performance, as they represent lower FLOPs and higher accuracy simultaneously.

Our proposed optimizable $\epsilon$ approach demonstrates more pronounced and consistent performance compared to both the baseline and the NEq method based on the ultimate optimizer. We attribute this improvement to the initial small value of $\epsilon$ during training, which enables the network to effectively learn the features. In general, as the number of training iterations increases, overfitting of certain features often occurs. However, the gradual increment of $\epsilon$ in our method helps mitigate the overfitting issue to some extent. This contributes to the enhanced and stable performance observed in our proposed approach.

\subsection{Experimental results}
\label{sec: Experimental Results}
We have chosen a selection of the most representative models currently available: VGG16, ResNet-18/32/56, MobileNet-v2, and Swin-T.
Among them, MobileNet-v2 and Swin-T use a pre-trained model, on ImageNet1K.

In all the experiments based on the SGD scheduler, we set the initial learning rate $\alpha$ to 0.1 (initialized to 0.01 for MobileNet-v2 and Swin-T) and reduced it by a factor of 10 after 100 and 150 epochs (after 30 and 60 for MobileNet-v2 and Swin-T). The momentum $\mu$ was set to 0.9, weight decay to 5 × $10^{-4}$, and the threshold $\epsilon$ to 0.001, with a total of 250 epochs (90 epochs for MobileNet-v2 and Swin-T). The parameter settings for the ultimate optimizer-based experiments were largely the same as those for the SGD-based experiments, with the key difference being the use of a hyper-optimizer to optimize the initial learning rate instead of the scheduler. $\eta_\alpha$ played a crucial role in the ultimate optimizer as it was used to optimize the learning rate. After conducting tests, we identified a suitable value of $\eta_\alpha$ for each architecture, ensuring optimal performance. For SCoTTi, we primarily adopted the hyperparameters of the ultimate optimizer, except that $\epsilon$ was initialized to 0, and $\eta_\epsilon$ was set to half of $\eta_\alpha$.

The experimental results are presented in Tab.~\ref{tab:results}. Our proposed approach is versatile, allowing for substantial FLOPs reduction across various datasets and architectures, while also slightly improving accuracy.

\section{Conclusion}
\label{sec:conclusion}

This study has presented the SCoTTi, an adaptive training framework for on-device training. The primary objective of SCoTTi was to reduce the number of FLOPs without compromising the model's accuracy. Remarkably, the results showed that in the majority of experimental cases, SCoTTi not only achieved state-of-the-art performance but also demonstrated the potential to enhance model accuracy in some of the most popular architectures on downstream tasks. Future works for the SCoTTi framework include investigating its adaptability to diverse domains such as natural language processing and audio processing, tailoring it to various hardware platforms for optimal on-device training, and conducting large-scale deployment experiments to assess its viability for real-world applications.

\section*{Acknowledgements} 
The research leading to these results has received funding from the project titled ``PC2-NF-NAI'' in the frame of the program ``PEPR 5G et Réseaux du futur'', and by Hi!PARIS Center on Data Analytics and Artificial Intelligence.

{\small
\bibliographystyle{unsrt}
\bibliography{main}
}

\end{document}